\documentclass[runningheads]{llncs}

\usepackage{eccv}

\usepackage{eccvabbrv}

\usepackage{graphicx}
\usepackage{booktabs}

\usepackage[accsupp]{axessibility}  

\usepackage[pagebackref,breaklinks,colorlinks,citecolor=eccvblue]{hyperref}

\usepackage{orcidlink}

\newcommand\customparagraph[1]{\vspace{0.4em}\noindent\textbf{#1.}}
\usepackage{multirow}
\usepackage{colortbl} 
\usepackage{xcolor}
\usepackage{wrapfig}

\begin{document}

\title{Few-shot NeRF by Adaptive Rendering Loss Regularization} 

\titlerunning{Few-shot NeRF by Adaptive Rendering Loss Regularization}

\author{Qingshan Xu\inst{1}\orcidlink{0000-0003-0405-3962} \and
Xuanyu Yi\inst{1} \and
Jianyao Xu\inst{2} \and
Wenbing Tao\inst{2}\orcidlink{0000-0003-3284-864X} \and
Yew-Soon Ong\inst{1,3}\orcidlink{0000-0002-4480-169X} \and
Hanwang Zhang\inst{1}\orcidlink{0000-0001-7374-8739}
}

\authorrunning{Q.~Xu et al.}

\institute{CCDS, Nanyang Technological University, Singapore \and
School of AIA, Huazhong University of Science and Technology, Wuhan, China \and
Centre for Frontier AI Research, IHPC, A*STAR, Singapore}

\maketitle

\begin{abstract}
Novel view synthesis with sparse inputs poses great challenges to Neural Radiance Field (NeRF). Recent works demonstrate that the frequency regularization of Positional Encoding (PE) can achieve promising results for few-shot NeRF. In this work, we reveal that there exists an inconsistency between the frequency regularization of PE and rendering loss. This prevents few-shot NeRF from synthesizing higher-quality novel views. To mitigate this inconsistency, we propose Adaptive Rendering loss regularization for few-shot NeRF, dubbed AR-NeRF. Specifically, we present a two-phase rendering supervision and an adaptive rendering loss weight learning strategy to align the frequency relationship between PE and 2D-pixel supervision. In this way, AR-NeRF can learn global structures better in the early training phase and adaptively learn local details throughout the training process. Extensive experiments show that our AR-NeRF achieves state-of-the-art performance on different datasets, including object-level and complex scenes. 
Our code will be available at \url{https://github.com/GhiXu/AR-NeRF}.
\keywords{Few-shot NeRF \and Adaptive rendering loss regularization \and Adaptive rendering loss weight learning}
\end{abstract}    
\section{Introduction}
\label{sec:intro}

Neural Radiance Field (NeRF)~\cite{mildenhall2021nerf} has gained tremendous popularity in synthesizing high-quality novel views of a scene. It represents the scene as a color and density field by a Multi-Layer Perceptron (MLP). Given many ground-truth 2D views of the scene, \eg, a dense input of 50--100 viewpoint images, NeRF optimizes the MLP representation by the following steps: 1) sampling a set of 3D points along camera rays\footnote[1]{A ray is a line emitted from a camera towards a pixel in that camera's 2D view.}, 2) encoding each 3D point into a vector by positional encoding (PE), 3) feeding the vector encoding into the MLP to output the color and density of the point, 4) integrating the colors and densities of the sampled points of each camera ray $\mathbf{y}$ into a 2D rendered color $\hat{\mathbf{c}}(\mathbf{y})$, and 5) minimizing the rendering loss $\mathcal{L}=\sum_{\mathbf{y}\in\mathcal{Y}}||\hat{\mathbf{c}}(\mathbf{y})-\mathbf{c}(\mathbf{y})||_2^2$ via gradient descent, where $\mathcal{Y}$ and $\mathbf{c}(\mathbf{y})$ denote the sampled rays and the ground-truth RGB supervision of the input image pixel, respectively. However, for many real-time applications, such as AR/VR and autonomous driving, it is usually impossible to collect many viewpoint images as dense input.

\begin{figure*}[t]
    \centering
    \includegraphics[width=\linewidth]{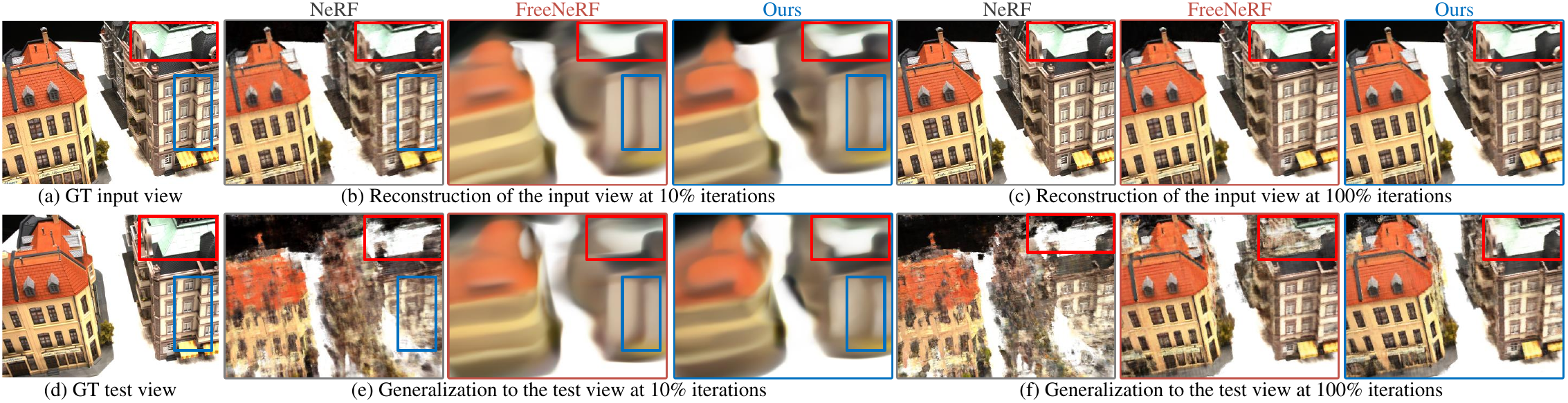}
    \caption{\textbf{Illustration of the relationship between Positional Encoding (PE) and rendering.} We show the reconstruction and generalization performance of different methods under 3 input views. For FreeNeRF~\cite{yang2023freenerf} and our method, AR-NeRF, 10\% iterations mean low-frequency PEs are enabled while 100\% iterations mean PEs with all frequencies are enabled. The red boxes show low-frequency global structures while the blue boxes show high-frequency local details. 
    }
    \label{fig:teaser}
\end{figure*}

When directly applying NeRF to sparse inputs, it is easily overfitted (the left of \cref{fig:teaser}(f)), \ie, the MLP optimization is especially sensitive to high-frequency information---local details like rich texture, but ignores low-frequency information---global structure like shape. Recently, several works~\cite{yang2023freenerf,truong2023sparf} have demonstrated that the overfitting is caused by PE in Step 2. The standard PEs used in NeRF contain encodings with different frequencies:
\begin{equation}
    PE(\mathbf{a}) = [\mathbf{a}, \eta_0(\mathbf{a}), \eta_1(\mathbf{a}), \dots, \eta_{K-1}(\mathbf{a})],
    \label{eq:pe}
\end{equation}
where $\eta_{k}(\mathbf{a}) = [\sin(2^{k} \pi \mathbf{a}), \cos(2^{k} \pi \mathbf{a})]$,  $K$ is a hyperparameter that controls the maximum encoded frequency, $\mathbf{a}$ is point coordinate or viewing direction. 
The high-frequency PEs (larger $k$) allow NeRF to quickly focus on optimizing high-frequency pixels, helping synthesize high-frequency details~\cite{mildenhall2021nerf,tancik2020fourier}. However, these works point out that, focusing on high-frequency pixel optimization too early prevents NeRF from fully learning low-frequency global structures in the sparse input setting. Therefore, the NeRF optimization for low-frequency global structures will be negatively affected by the high-frequency local details.

To alleviate the above problem, FreeNeRF~\cite{yang2023freenerf} and SPARF~\cite{truong2023sparf} regularize the frequency of PE. By starting from feeding low-frequency PEs and then gradually feeding high-frequency ones, these works stabilize the NeRF by learning global structures first, then polishing local details. This ``global first, local second'' mechanism has been verified in many reconstruction tasks~\cite{wang2021neus,fu2022geo,yi2024diffusion,xu2019multi,xu2022multi,ren2023hierarchical}. However, the frequency relationship between the MLP input and output is not necessarily true for the frequency relationship between the PEs and 2D pixels. As shown in the red box on the center of \cref{fig:teaser}(e), the low-frequency PEs cannot learn some low-frequency information well. To formally see this, we decompose the loss in Step 5 into:     
\begin{equation}
\mathcal{L}=\overbrace{\sum_{\mathbf{y}\in\mathcal{Y_\text{low}}}||\hat{\mathbf{c}}(\mathbf{y})-\mathbf{c}(\mathbf{y})||_2^2}^{\mathcal{L}_\text{low}}+\overbrace{\sum_{\mathbf{y}\in\mathcal{Y_\text{high}}}||\hat{\mathbf{c}}(\mathbf{y})-\mathbf{c}(\mathbf{y})||_2^2}^{\mathcal{L}_\text{high}},
\label{eq:loss}
\end{equation}
where $\mathcal{Y}_\text{low}$ and $\mathcal{Y}_\text{high}$ denote the rays from low-frequency and high-frequency pixel supervision, respectively. We see that, when only inputting low-frequency PEs, \eg, $[\mathbf{a}, \eta_0(\mathbf{a}), \eta_1(\mathbf{a})]$, the supervision from high-frequency pixels, $\{\mathbf{c}(\mathbf{y})|\mathbf{y}\in \mathcal{Y}_\text{high}\}$ is still provided. This supervision will interfere with the ability of low-frequency PEs to adequately learn low-frequency pixels during the global structure learning (More empirical analysis will be shown in \cref{sec:experiment}).  
This demonstrates the inconsistency between the frequency regularization of PE and rendering loss. 
In fact, an ideal way to address this problem is to align the frequency relationship between PE and pixel supervision, \ie, when the PE of a specific frequency, $\eta_{k}(\mathbf{a})$, is input, providing the pixels corresponding to the frequency as pixel supervision. However, this is challenging since the corresponding ground-truth pixel supervision of $\eta_{k}(\mathbf{a})$ is unknown.  

\begin{figure*}[t]
    \centering
    \includegraphics[width=\linewidth]{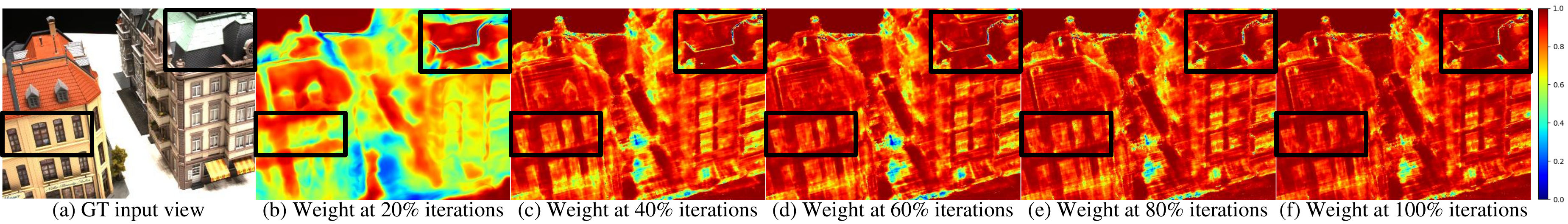}
    \caption{\textbf{Adaptive rendering loss weight learning at different iterations.} High-frequency PEs are gradually input during training. The larger the color value, the greater the weight. 
    Low-frequency global structures (\eg, roofs) have greater weights in the early training phase. In contrast, the weights of high-frequency local details (\eg, windows) are smaller. When high-frequency PEs are gradually input, the weights of global structures remain high while the weights of local details become greater and greater.
    }
    \label{fig:weight}
\end{figure*}

To this end, we propose Adaptive Rendering loss regularization for few-shot NeRF, dubbed AR-NeRF. Our key insight is to provide more low-frequency pixel supervision for \cref{eq:loss} in the early training phase, and adaptively learn the rendering loss weights for different pixel supervision when high-frequency PEs are gradually input.  
Specifically, besides the PE frequency regularization~\cite{yang2023freenerf, truong2023sparf}, we design:
\begin{itemize}
    \item A two-phase rendering supervision (\cref{sec:ar}). It uses blurred input views as pixel supervision in the early training phase, and then provides original input views as pixel supervision in the subsequent training phase. This reduces the high-frequency pixel supervision for \cref{eq:loss} in the early training phase, encouraging the low-frequency PEs to truly learn global structure information (the right side of \cref{fig:teaser}(e)). 
    \item An adaptive rendering loss weight learning strategy (\cref{sec:ar}) based on uncertainty learning~\cite{kendall2017uncertainties}. Uncertainty learning tends to learn easy examples with low uncertainty, and learn hard examples with high uncertainty. In fact, in the process of gradually inputting high-frequency PEs, the pixels corresponding to the frequencies of the input PEs are easy examples, while the pixels of other frequencies are hard examples. Therefore, when gradually inputting high-frequency PEs, we can use uncertainty to learn adaptive rendering loss weights for pixel supervision with different frequencies. As a result, more attention can be paid to low-frequency pixels in the early training phase, as shown in \cref{fig:weight}(b). Meanwhile, both high-frequency PEs and observed pixels will be gradually provided (\cref{fig:weight}(c) and (d)), local detail information can be gradually learned  (the right of \cref{fig:teaser}(f)).
\end{itemize}
In this way, 
our method can focus more on learning global structures in the early training phase, and adaptively learn local details without destroying global structures. 
Note that, our work is orthogonal to other works introducing external modules or obvious computational cost~\cite{niemeyer2022regnerf,wang2023sparsenerf,kwak2023geconerf,sun2024global}.

Overall, our contributions are summarized as follows:
\begin{itemize}
    \item We reveal the inconsistency between the frequency regularization of PE and rendering loss. This prevents few-shot NeRF from synthesizing higher quality novel views.
    \item To mitigate the inconsistency, adaptive rendering loss regularization is proposed to align the frequency relationship between PE and pixel supervision. This allows our method to learn better global structures first and adaptively learn local details throughout the training process.
    \item Our approach achieves state-of-the-art performance across different datasets. Especially, our approach does not introduce external modules or significant computational cost, making it a friendly solution in practice.
\end{itemize}
\section{Related Work}
\label{sec:related}

\customparagraph{Neural Radiance Fields}  
NeRF~\cite{mildenhall2021nerf} maps 5D coordinates to a color and geometry field to represent 3D scenes. 
Due to its simple concept and impressive performance, it has shown great success in a variety of applications, such as novel view synthesis~\cite{barron2021mip,barron2022mip,verbin2022ref,martin2021nerf}, reconstruction~\cite{yariv2021volume,wang2021neus,fu2022geo,xu2023pr,su2024psdf,wang2024pgahum}, 3D generation~\cite{chan2021pi,poole2022dreamfusion,lin2023magic3d,yi2024diffusion,esposito2024geogen} and so on. 
Although NeRF and its follow-ups have made significant progress, they usually require a dense set of input images. Real-world applications sometimes can only capture limited input views, making these vanilla methods degenerate a lot in practice. Our work focuses on improving the generalization performance on unseen viewpoints for NeRF with sparse inputs.   

\customparagraph{Few-shot Neural Radiance Fields} Few-shot NeRF faces great challenges due to limited rendering supervision. Therefore, many methods attempt to address this challenge by utilizing extra information. These methods can be categorized into two groups: pre-training and regularization method.

The pre-training methods~\cite{yu2021pixelnerf,chen2021mvsnerf,johari2022geonerf} learn generalization models by training on a large-scale multi-view dataset. By conditioning the models on input image features, the models learn extensive prior knowledge. In test scenes, these methods need to further fine-tune their models to adapt to specific scenes. The pre-training methods require expensive data collection and are apt to performance degeneration due to domain gap. Another line of research regularizes color and geometry fields by introducing extra supervision. The extra supervision includes depth supervision~\cite{deng2022depth,roessle2022dense,wang2023sparsenerf,sun2024global,uy2023scade}, normalization-flow~\cite{niemeyer2022regnerf}, semantic consistency~\cite{jain2021putting} and learned priors from denoising diffusion models~\cite{wynn2023diffusionerf,wu2024reconfusion}. Some methods~\cite{niemeyer2022regnerf,chen2022geoaug,bai2023self,kwak2023geconerf,kim2022infonerf,seo2023flipnerf} leverage data augmentation or pseudo views to reduce the risk of overfitting. Although these methods have achieved promising results, they make the few-shot NeRF training become inefficient by introducing external modules or extra computational cost. 

Unlike the aforementioned methods, some recent works~\cite{yang2023freenerf,seo2023mixnerf,truong2023sparf} demonstrate that it is possible for few-shot NeRF to enhance generalization to unseen viewpoints without external modules or obvious computational cost. FreeNeRF~\cite{yang2023freenerf} and SPARF~\cite{truong2023sparf} show that using a simple frequency regularization of PE largely prevents overfitting. 
On this basis, we reveal the inconsistency between the frequency regularization of PE and rendering loss and propose adaptive rendering loss regularization.

\customparagraph{Uncertainty in Neural Radiance Fields} Uncertainty learning has been used by some NeRF methods~\cite{shen2021stochastic,shen2022conditional,pan2022activenerf} to quantify the reliability of final synthesized views under \emph{dense inputs} and \emph{fixed PE} settings. NeRF-W~\cite{martin2021nerf} introduces uncertainty to model transparent objects in the scene. Different from these works, we use the uncertainty learning~\cite{kendall2017uncertainties} to align the frequency relationship between PE and pixel supervision for few-shot NeRF. 
\section{Method}
\label{sec:method}

\begin{figure*}[t]
    \centering
    \includegraphics[width=\linewidth]{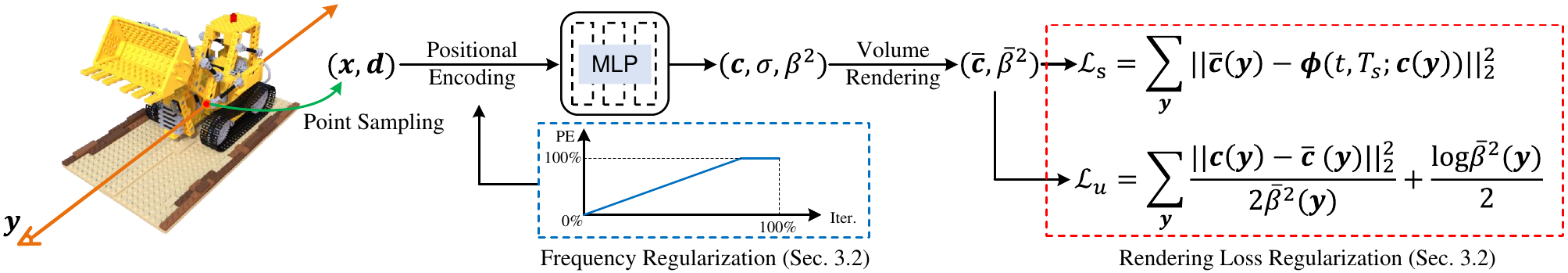}
    \caption{\textbf{Overview of AR-NeRF.} Our method follows the training pipeline described in \cref{sec:intro}. Besides the frequency regularization of PE, we propose adaptive rendering loss regularization. This aligns the frequency relationship between PE and pixel supervision. See \cref{sec:method} for more details.}
    \label{fig:overview}
\end{figure*}

\subsection{Preliminaries}

NeRF~\cite{mildenhall2021nerf} represents a 3D scene as a color field and density field by using an MLP. After sampling a set of 3D points in Step 1, PE encodes a 3D position $\mathbf{x}\in\mathbb{R}^3$ and a viewing direction $\mathbf{d}\in\mathbb{S}^2$ into high-dimension vectors in Step 2 by \cref{eq:pe}. In Step 3, the vector encodings are further fed into the MLP to output the corresponding color $\mathbf{c}\in[0,1]^3$ and density $\sigma\in\mathbb{R}^{+}$:
\begin{equation}
    F_\theta(PE(\mathbf{x}), PE(\mathbf{d})) = (\mathbf{c}, \sigma),
\end{equation}
where $F$ is an MLP with parameters $\theta$.

In order to render a pixel on an image in Step 4, a ray $\mathbf{y}(t)=\mathbf{o}+t\mathbf{d}$ is emitted from the camera center $\mathbf{o}$ through the pixel along the direction $\mathbf{d}$. Following the volume rendering~\cite{max1995optical}, the pixel is rendered by alpha compositing the colors and densities along the ray using the quadrature rule:
\begin{equation}
\begin{split}
    \hat{\mathbf{c}}(\mathbf{y}) &= \sum_{i=1}^{N} w_i \mathbf{c}_{i} = \sum_{i=1}^{N} T_{i} \alpha_i \mathbf{c}_{i}, \\
    \text{with} \quad T_{i} &= \exp(-\sum_{j=1}^{i-1}\sigma_j \delta_j), ~\alpha_i = 1 - \exp(-\sigma_{i} \delta_{i}),
\end{split}
\label{eq:render}
\end{equation}
where $N$ is the total sample number along the ray, $w_i$ denotes the weight of each point on the ray, $T_i$ is the transmittance, $\alpha_i$ is the opacity and $\delta_i$ is the distance between the adjacent sample points. 
In Step 5, NeRF minimizes the rendering loss $\mathcal{L}$, \ie, the mean squared error (MSE) between the rendered color $\hat{\mathbf{c}}(\mathbf{y})$ and ground truth pixel color supervision $\mathbf{c}(\mathbf{y})$ to optimize the MLP representation.

\subsection{Adaptive Rendering Loss Regularization}
\label{sec:ar}

In the sparse input setting, the goal of AR-NeRF is to focus more on learning low-frequency global structures in the early training phase, and adaptively learn high-frequency local details throughout the training process. 
We achieve this goal from two perspectives: the input to MLP, \ie, PE, and the rendering supervision of MLP, \ie,  rendering loss. In terms of the input to MLP, we adopt the frequency regularization of PE to gradually input high-frequency PEs~\cite{yang2023freenerf,truong2023sparf}. In terms of the rendering supervision of MLP, we propose adaptive rendering loss regularization. Furthermore, this rendering loss regularization contains two designs: two-phase rendering supervision and adaptive rendering loss weight learning. The overview of AR-NeRF is shown in \cref{fig:overview}.

\customparagraph{Frequency regularization of PE} As pointed out in \cite{yang2023freenerf,truong2023sparf}, the standard PE (\cref{eq:pe}) used in NeRF prevents few-shot NeRF from fully learning low-frequency global information. To alleviate this, we regularize the frequency of PE~\cite{yang2023freenerf,truong2023sparf}. Given a PE of length $L+3$, a linearly increasing frequency mask $\mathbf{m}$ is applied to PE as follows:
\begin{equation}
    PE'(t, T;\mathbf{a}) = PE(\mathbf{a}) \odot \mathbf{m}(t, T, L),
\end{equation}
\begin{equation}
\begin{split}
    m_i(t,T,L)= \left \{
\begin{array}{ll}
    1, & \text{if}~i\leq \frac{t\cdot L}{T}+3 \\
    \frac{t\cdot L}{T}-\lfloor\frac{t\cdot L}{T}\rfloor, & \text{if}~\frac{t\cdot L}{T}+3<i\leq\frac{t\cdot L}{T}+6 \\
    0, & \text{if}~i>\frac{t\cdot L}{T}+6
\end{array}
\right.
\end{split}
\end{equation}
where $m_i(t,T,L)$ denotes the $i$-th bit value of $\mathbf{m}(t,T,L)$; $t$ and $T$ are the current iteration number and the maximum iteration number of frequency regularization. This makes the input to MLP begin with raw inputs $(\mathbf{x}, \mathbf{d})$ and gradually inputs PE from low-frequency to high-frequency. 

The frequency regularization reduces the sensitivity of NeRF optimization to high-frequency 2D pixels in the early training phase. However, in the progress of gradually inputting PEs, since the rendering loss always contains high-frequency 2D pixel supervision (\cref{eq:loss}),  
this supervision will prevent low-frequency PEs from adequately learning low-frequency information during the global structure learning. 
This leads to the inconsistency between the frequency regularization of PE and rendering loss. Therefore, we propose adaptive rendering loss regularization to align the frequency relationship between PE and 2D pixel supervision.

\customparagraph{Two-phase rendering supervision} To mitigate the above inconsistency, we design a two-phase rendering supervision to provide low-frequency 2D pixel supervision as much as possible for low-frequency PEs in the early training phase. In the subsequent phase, we use raw pixel supervision from ground-truth input views to optimize NeRF. Concretely, we define a two-phase transformation as:
\begin{equation}
\begin{split}
\Phi(t,T_s;\cdot) = \left \{
\begin{array}{ll}
        G(\cdot), & \text{if}\quad t < T_s  \\
        \mathrm{I}, & \text{if}\quad t\geq T_s \\
\end{array}
\right.
\end{split}\label{eq:blur}
\end{equation}
where $G(\cdot)$ is the Gaussian blur function, $\mathrm{I}$ is an Identity mapping function and $T_s$ is the maximum iteration number of early training phase. By applying the above transformation to all ground-truth input views, our rendering loss becomes as follows:
\begin{equation}
    \mathcal{L}_s = \sum_{\mathbf{y}\in\mathcal{Y}}||\hat{\mathbf{c}}(\mathbf{y})-\Phi(t, T_s;\mathbf{c}(\mathbf{y}))||_2^2.
\end{equation}
In this way, the two-phase rendering supervision provides more low-frequency 2D pixel supervision for \cref{eq:loss} in the early training phase. This can better guide low-frequency PEs to learn low-frequency 2D pixels. As a result, our method can learn global structures better in the early training phase, thus building a better foundation for subsequent local detail optimization.

\customparagraph{Adaptive rendering loss weight learning}  
The  two-phase rendering supervision mainly works for some specific low-frequency PEs (\eg, $[\mathbf{a}, \eta_0(\mathbf{a}), \eta_1(\mathbf{a})]$) in the early training phase.
Since PEs contain the encodings with different frequencies (\cref{eq:pe}), it is better to adaptively adjust the rendering loss weight for 2D pixel supervision with different frequencies. However, it is challenging to associate PE frequency with pixel supervision frequency. 
Inspired by the uncertainty learning~\cite{kendall2017uncertainties}, we design an adaptive rendering loss weight learning strategy. 

The uncertainty learning~\cite{kendall2017uncertainties} tends to learn easy examples with low uncertainty and to learn hard examples with high uncertainty. This uncertainty can be learned by deep neural networks in a self-supervised way. When we input PEs of specific frequencies, NeRF will tend to learn 2D pixels corresponding to these frequencies better. For 2D pixels with other frequencies, it will be a bit hard for NeRF to learn these pixels. That is, the former pixels, in fact, are easy examples while the latter ones are hard examples. Therefore, when gradually inputting high-frequency PEs, we can use the uncertainty to learn adaptive rendering loss weights for pixel supervision with different frequencies. 

To this aim, for a 3D position $\mathbf{x}$ with viewing direction $\mathbf{d}$, the corresponding color field is modeled as a Gaussian distribution with mean $\mathbf{c}$ and variance $\beta^2$. The corresponding density field is still parameterized as $\sigma$. Then, we have: 
\begin{equation}
    F_\theta(PE'(t,T;\mathbf{x}), PE'(t,T;\mathbf{d})) = (\mathbf{c}, \beta^2, \sigma).
\end{equation}
The variance measures uncertainty~\cite{kendall2017uncertainties}. Similarly, the rendered color mean $\bar{\mathbf{c}}(\mathbf{y})$ can be computed using \cref{eq:render}. We assume the color distribution of each 3D point is independent, the rendered color variance can be computed as:
\begin{equation}
    \bar{\beta}^2(\mathbf{y}) = \sum_{i=1}^{N} w_i^2 \beta^2.
\end{equation}
In this way, the rendered color $\hat{\mathbf{c}}(\mathbf{y})$ for ray $\mathbf{y}$ still approximately follows a Gaussian distribution, $\mathcal{N}(\bar{\mathbf{c}}(\mathbf{y}), \bar{\beta}^2(\mathbf{y}))$. By taking the negative log likelihood of this distribution, our optimization objective is defined as:
\begin{equation}
    -\log p(\mathbf{c}(\mathbf{y})|\mathbf{y}) \propto \frac{||\mathbf{c}(\mathbf{y}) - \bar{\mathbf{c}}(\mathbf{y})||_2^2}{2\bar{\beta}^2(\mathbf{y})} + \frac{\log \bar{\beta}^2(\mathbf{y})}{2}.
\end{equation}
In this way, we can define an adaptive rendering loss as:
\begin{equation}
    \mathcal{L}_u = \sum_{\mathbf{y}\in\mathcal{Y}} \frac{||\mathbf{c}(\mathbf{y}) - \bar{\mathbf{c}}(\mathbf{y})||_2^2}{2\bar{\beta}^2(\mathbf{y})} + \frac{\log \bar{\beta}^2(\mathbf{y})}{2}. 
\end{equation}
In our scenario, $1/\bar{\beta}^2(\mathbf{y})$ is cast as an adaptive learnable weight, which can be adjusted adaptively with the gradual input of high-frequency PEs (\cref{fig:weight}). 
It indicates the frequency relationship between PE and pixel supervision.

The above designs allow our method to align the frequency relationship between PE and pixel supervision. Therefore, our method is able to focus on learning low-frequency global structures in the early training phase (\cref{fig:weight}(b)), and adaptively learn high-frequency local details.

\subsection{Ray Density Regularization}

The sparse inputs lead to limited camera ray sampling. Therefore, the rendered colors of these rays cannot completely constrain the entire scene space. This may cause floating artifacts in synthesized images. To alleviate this, we further propose a ray density regularization loss to impose constraints on the sampled points. Following ~\cite{kim2022infonerf}, we use the opacity $\alpha_i$ to define the ray density as:
\begin{equation}
    p_i(\mathbf{y}) = \frac{\alpha_i}{\sum_j \alpha_j}.
\end{equation}
This considers the irregular sampling interval $\delta_i$. By applying the Emptiness loss~\cite{wang2023score} to the ray density, our ray density regularization loss is computed as:
\begin{equation}
    \mathcal{L}_r = \sum_{\mathbf{y}\in\mathcal{Y}} \frac{1}{N} \sum_{i=1}^N \log(1 + s \cdot p_i(\mathbf{y})),
    \label{eq:rdr}
\end{equation}
where $s$ is a constant to control the steepness of the loss function near 0. This loss encourages sparsity of 3D space by imposing more penalties on small ray densities. In this way, this regularization reduces floating artifacts.

\subsection{Loss Function}

During training, we regularize not only the rendering loss, but also the ray density. Our total loss is:
\begin{equation}
    \mathcal{L}_\text{total} = \mathcal{L}_s + \lambda_u \mathcal{L}_u  + \lambda_r \mathcal{L}_r + \lambda_o \mathcal{L}_o,
    \label{eq:total_loss}
\end{equation}
where $\mathcal{L}_o$ is an occlusion regularization loss to penalize the density fields near the camera~\cite{yang2023freenerf}; $\lambda_u$, $\lambda_r$ and $\lambda_o$ are balancing terms for the losses.

\section{Experiments}
\label{sec:experiment}

\subsection{Experimental Details}

\customparagraph{Datasets \& metrics} We evaluate AR-NeRF on two datasets under the sparse input setting: DTU dataset~\cite{jensen2014large} and LLFF dataset~\cite{mildenhall2019local}. The DTU dataset consists of images capturing objects located on a white table with a black background.  
The LLFF dataset contains real forward-facing complex scenes. 
Following~\cite{niemeyer2022regnerf}, we evaluate our AR-NeRF in the 3, 6, and 9 input-view settings on DTU and LLFF. 

To evaluate the quality of synthesized images, several quantitative metrics are adopted, including the
peak signal-to-noise ratio (PSNR), structural similarity index (SSIM)~\cite{wang2004image} and learned perceptual image patch similarity (LPIPS)~\cite{zhang2018unreasonable}. Following~\cite{niemeyer2022regnerf}, the geometric mean of $\mathrm{MSE}=10^{-\mathrm{PSNR}/{10}}$, $\sqrt{1-\mathrm{SSIM}}$, and LPIPS is also reported as an "Average" score. On DTU dataset, we use the object masks to remove the background when computing these metrics. This avoids biasing towards the background when using full-image evaluation~\cite{niemeyer2022regnerf,yu2021pixelnerf}.

\begin{table*}
    \caption{
    \textbf{Quantitative Comparison on DTU.}
    For $3$ input views, our method achieves the best results in all metrics. 
    For $6$ and $9$ input views, our model achieves comparable or state-of-the-art performance in most metrics. 
    The best, second-best, and third-best entries are marked in red, orange, and yellow, respectively.
    }
    \resizebox{\linewidth}{!}{
\begin{tabular}{l|c|ccc|ccc|ccc|ccc}
\toprule
  & \multirow{2}{*}{Setting} &  \multicolumn{3}{c}{PSNR $\uparrow$} & \multicolumn{3}{c}{SSIM $\uparrow$} & \multicolumn{3}{c}{LPIPS $\downarrow$} & \multicolumn{3}{c}{Average $\downarrow$}  \\
  &  & 3-view & 6-view & 9-view  & 3-view & 6-view & 9-view  & 3-view & 6-view & 9-view  & 3-view & 6-view & 9-view \\ \midrule
SRF~\cite{chibane2021stereo} & \multirow{3}{*}{Trained on DTU} & 15.32 & 17.54 & 18.35 & 0.671 & 0.730 & 0.752 & 0.304 & 0.250 & 0.232 & 0.171 & 0.132 & 0.120 \\
PixelNeRF~\cite{yu2021pixelnerf} &  & 16.82 & 19.11 & 20.40 & 0.695 & 0.745 & 0.768 & 0.270 & 0.232 & 0.220 & 0.147 & 0.115 & 0.100 \\
MVSNeRF~\cite{chen2021mvsnerf} &  &  18.63 &  20.70 & 22.40 &  \cellcolor{yellow!25}0.769 &  0.823 &  0.853 &  0.197 &  0.156 &  0.135 &  0.113 &  0.088 & 0.068 \\
\midrule 
SRF ft~\cite{chibane2021stereo} & \multirow{3}{*}{\shortstack{Trained on DTU\\and\\Optimized per Scene}} & 15.68 & 18.87 & 20.75 & 0.698 & 0.757 & 0.785 & 0.281 & 0.225 & 0.205 & 0.162 & 0.114 & 0.093 \\
PixelNeRF ft~\cite{yu2021pixelnerf} &  &  \cellcolor{yellow!25}18.95 & 20.56 & 21.83 & 0.710 & 0.753 & 0.781 & 0.269 & 0.223 & 0.203 & 0.125 & 0.104 & 0.090 \\
MVSNeRF ft~\cite{chen2021mvsnerf} &  & 18.54 & 20.49 & 22.22 &  \cellcolor{yellow!25}0.769 &  0.822 &  0.853 &  0.197 &  0.155 &  0.135 &  0.113 &  0.089 & 0.069 \\
\midrule 
mip-NeRF~\cite{barron2021mip} & \multirow{6}{*}{Optimized per Scene} & 8.68 & 16.54 &  \cellcolor{yellow!25}23.58 & 0.571 & 0.741 &   0.879 & 0.353 & 0.198 &  \cellcolor{orange!25}0.092 & 0.323 & 0.148 &  0.056 \\
DietNeRF~\cite{jain2021putting} &  & 11.85 &  20.63 &  23.83 & 0.633 & 0.778 & 0.823 & 0.314 & 0.201 & 0.173 & 0.243 & 0.101 &  0.068 \\
RegNeRF~\cite{niemeyer2022regnerf} &  & 18.89 &  22.20 &  24.93 &  0.745 &  \cellcolor{orange!25}0.841 &  \cellcolor{red!25}0.884 &  \cellcolor{yellow!25}0.190 &  \cellcolor{red!25}0.117 &  \cellcolor{red!25}0.089 &  \cellcolor{yellow!25}0.112 &  \cellcolor{yellow!25}0.071 &  \cellcolor{red!25}0.047 \\
MixNeRF~\cite{seo2023mixnerf} & & \cellcolor{yellow!25}18.95 & \cellcolor{yellow!25}22.30 & \cellcolor{yellow!25}25.03 & 0.744 & \cellcolor{yellow!25}0.835 & 0.879 & 0.228 & \cellcolor{yellow!25}0.139 & \cellcolor{yellow!25}0.105 & 0.118 & 0.073 & \cellcolor{yellow!25}0.050 \\
FreeNeRF~\cite{yang2023freenerf} & & \cellcolor{orange!25}19.92 & \cellcolor{orange!25}23.14 & \cellcolor{orange!25}25.60 & \cellcolor{orange!25}0.787 & \cellcolor{red!25}0.844 & \cellcolor{orange!25}0.883 & \cellcolor{red!25}0.182 & \cellcolor{orange!25}0.137 & 0.109 & \cellcolor{orange!25}0.098 & \cellcolor{red!25}0.068 & \cellcolor{orange!25}0.048 \\
\textbf{AR-NeRF (Ours)} &  & \cellcolor{red!25}20.36 &  \cellcolor{red!25}23.17 &  \cellcolor{red!25}25.62 &  \cellcolor{red!25}0.788 &  \cellcolor{orange!25}0.841 &  \cellcolor{yellow!25}0.881 &  \cellcolor{orange!25}0.187 &  0.143 &  0.114 &  \cellcolor{red!25}0.095 &  \cellcolor{orange!25}0.070 &  \cellcolor{orange!25}0.048 \\
\bottomrule
\end{tabular}}
    \label{tab:dtu}
\end{table*}

\begin{figure*}
    \centering
    \includegraphics[width=\linewidth]{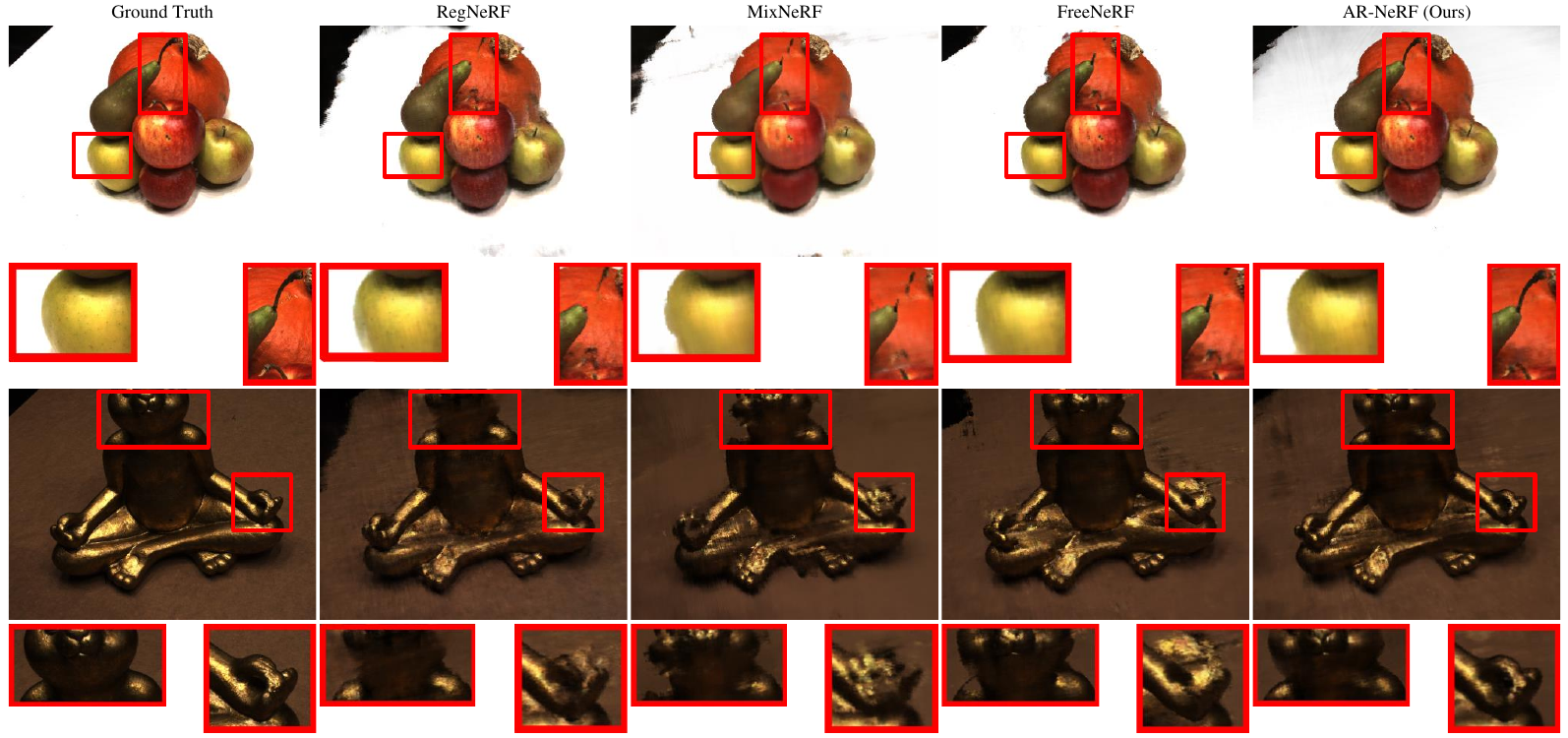}
    \caption{\textbf{Qualitative comparison on DTU.} We show novel views rendered by different methods in 3 input-view setting.}
    \label{fig:dtu_comp}
\end{figure*}

\customparagraph{Implementations} We implement AR-NeRF based on the JAX codebase~\cite{bradbury2018jax}. The Adam optimizer~\cite{kingma2014adam} is used for the learning of AR-NeRF. The exponential decay and warm up are applied for the learning rate. The batch size is set to 4096. We train our AR-NeRF on one NVIDIA RTX 3090. For the balancing terms used in \cref{eq:total_loss}, $\lambda_u=\lambda_o=0.01$, $\lambda_r$ is initialized as $10^{-5}$ and then linearly increased to $10^{-3}$ in 512 iterations. We set $s=10.0$ for \cref{eq:rdr}. The Gaussian blur kernel size for \cref{eq:blur} is set to 3.  We follow FreeNeRF~\cite{yang2023freenerf} to set $T$. On the DTU and LLFF datasets, $T_s$ is set to $10000$, $15000$ and $16000$ for the 3, 6, and 9 input-view setting, respectively. See supplementary for more analysis on these parameters.

\begin{table*}
    \caption{
    \textbf{Quantitative Comparison on LLFF.}
    Our method achieves the best results in most metrics under different input-view settings.  
    The best, second-best, and third-best entries are marked in red, orange, and yellow, respectively. 
    }
    \resizebox{\linewidth}{!}{
\begin{tabular}{l|c|ccc|ccc|ccc|ccc}
\toprule
  & \multirow{2}{*}{Setting} &  \multicolumn{3}{c}{PSNR $\uparrow$} & \multicolumn{3}{c}{SSIM $\uparrow$} & \multicolumn{3}{c}{LPIPS $\downarrow$} & \multicolumn{3}{c}{Average $\downarrow$}  \\
  &  & 3-view & 6-view & 9-view  & 3-view & 6-view & 9-view  & 3-view & 6-view & 9-view  & 3-view & 6-view & 9-view \\ \midrule
SRF~\cite{chibane2021stereo} & \multirow{3}{*}{Trained on DTU} & 12.34 & 13.10 & 13.00 & 0.250 & 0.293 & 0.297 & 0.591 & 0.594 & 0.605 & 0.313 & 0.293 & 0.296 \\
PixelNeRF~\cite{yu2021pixelnerf} &  & 7.93 & 8.74 & 8.61 & 0.272 & 0.280 & 0.274 & 0.682 & 0.676 & 0.665 & 0.461 & 0.433 & 0.432 \\
MVSNeRF~\cite{chen2021mvsnerf} &  & 17.25 & 19.79 & 20.47 & 0.557 & 0.656 & 0.689 & 0.356 & 0.269 & 0.242 & 0.171 & 0.125 & 0.111 \\
\midrule 
SRF ft~\cite{chibane2021stereo} & \multirow{3}{*}{\shortstack{Trained on DTU\\and\\Optimized per Scene}} & 17.07 & 16.75 & 17.39 & 0.436 & 0.438 & 0.465 & 0.529 & 0.521 & 0.503 & 0.203 & 0.207 & 0.193 \\
PixelNeRF ft~\cite{yu2021pixelnerf} &  & 16.17 & 17.03 & 18.92 & 0.438 & 0.473 & 0.535 & 0.512 & 0.477 & 0.430 & 0.217 & 0.196 & 0.163 \\
MVSNeRF ft~\cite{chen2021mvsnerf} &  & 17.88 & 19.99 & 20.47 & 0.584 & 0.660 & 0.695 & 0.327 & 0.264 & 0.244 & 0.157 & 0.122 & 0.111 \\
\midrule 
mip-NeRF~\cite{barron2021mip} & \multirow{6}{*}{Optimized per Scene} & 14.62 & 20.87 & 24.26 & 0.351 & 0.692 & 0.805 & 0.495 & 0.255 & 0.172 & 0.246 & 0.114 & 0.073 \\
DietNeRF~\cite{jain2021putting} &  & 14.94 & 21.75 & 24.28 & 0.370 & 0.717 & 0.801 & 0.496 & 0.248 &0.183 & 0.240 & 0.105 & 0.073 \\
RegNeRF~\cite{niemeyer2022regnerf} &  & 19.08 & 23.10 & 24.86 & 0.587 & 0.760 & 0.820 & 0.336 & 0.206 & 0.161 & 0.146 & 0.086 & 0.067 \\
MixNeRF~\cite{seo2023mixnerf} & & \cellcolor{yellow!25}19.27 & \cellcolor{orange!25}23.76 & \cellcolor{orange!25}25.20 & \cellcolor{orange!25}0.629 & \cellcolor{red!25}0.791 & \cellcolor{orange!25}0.833 & \cellcolor{orange!25}0.301 & \cellcolor{orange!25}0.185 & \cellcolor{orange!25}0.153 & \cellcolor{yellow!25}0.135 & \cellcolor{orange!25}0.078 & \cellcolor{orange!25}0.063 \\
FreeNeRF~\cite{yang2023freenerf} & & \cellcolor{orange!25}19.63 & \cellcolor{yellow!25}23.73 & \cellcolor{yellow!25}25.13 & \cellcolor{yellow!25}0.612 & \cellcolor{yellow!25}0.779 & \cellcolor{yellow!25}0.827 & \cellcolor{yellow!25}0.308 & \cellcolor{yellow!25}0.195 & \cellcolor{yellow!25}0.160 & \cellcolor{yellow!25}0.134 & \cellcolor{yellow!25}0.079 & \cellcolor{yellow!25}0.064 \\
\textbf{AR-NeRF (Ours)} &  & \cellcolor{red!25}19.90 & \cellcolor{red!25}23.84 & \cellcolor{red!25}25.23 & \cellcolor{red!25}0.635 & \cellcolor{orange!25}0.785 & \cellcolor{red!25}0.836 & \cellcolor{red!25}0.283 & \cellcolor{red!25}0.182 & \cellcolor{red!25}0.144 & \cellcolor{red!25}0.126 & \cellcolor{red!25}0.077 & \cellcolor{red!25}0.061 \\
\bottomrule
\end{tabular}}
    \label{tab:llff}
\end{table*}

\begin{figure*}
    \centering
    \includegraphics[width=\linewidth]{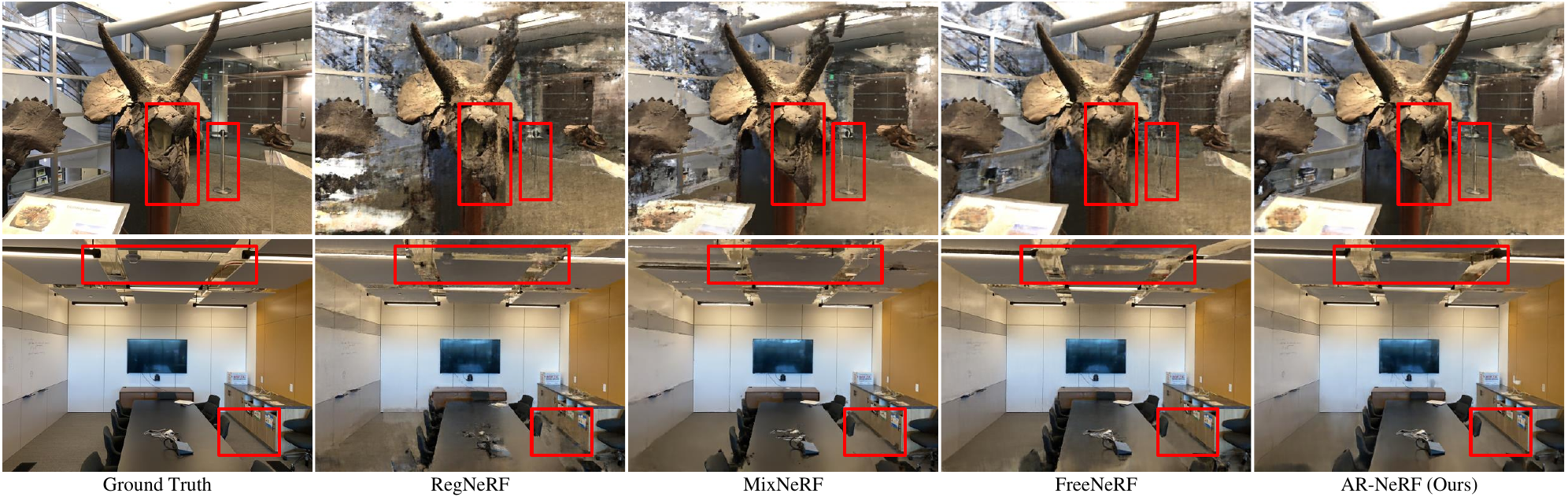}
    \caption{\textbf{Qualitative comparison on LLFF.} We show novel views rendered by different methods in 3 input-view setting.}
    \label{fig:llff_comp}
\end{figure*}

\customparagraph{Baselines} We compare our AR-NeRF against the state-of-the-art pre-training methods and regularization methods as well as the vanilla mip-NeRF~\cite{barron2021mip}. The pre-training methods include PixelNeRF~\cite{yu2021pixelnerf}, Stereo Radiance Fields (SRF)~\cite{chibane2021stereo}, and MVSNeRF~\cite{chen2021mvsnerf}. For the regularization methods, DietNeRF~\cite{jain2021putting}, RegNeRF~\cite{niemeyer2022regnerf}, MixNeRF\footnote[2]{MixNeRF uses AlexNet to evaluate LPIPS while other methods use VGG. For fair comparisons, we use its provided models to retest its performance by VGG.}~\cite{seo2023mixnerf} and FreeNeRF\footnote[3]{We retest the performance of FreeNeRF on DTU and LLFF by its provided models, and find that the results are slightly worse than the values reported in its paper.}~\cite{yang2023freenerf} are compared.

\subsection{Comparison}

\customparagraph{DTU dataset} 
\Cref{tab:dtu} shows the quantitative results on this dataset. Although the pre-training methods (SRF~\cite{chibane2021stereo}, PixelNeRF~\cite{yu2021pixelnerf} and MVSNeRF~\cite{chen2021mvsnerf}) are trained on this dataset, their performance underperforms ours in all settings. Compared with the regularization methods, our method achieves comparable or better results in most metrics under different input-view settings. Remarkably, our method achieves the best results in all metrics for 3 input views. This is because our method can learn global structures first, then gradually learn local details. \cref{fig:dtu_comp} shows the qualitative results in this setting. The object outlines, \eg, the statue head and hand, cannot be recovered well for RegNeRF, MixNeRF and FreeNeRF. Thus, they cannot reconstruct local details well due to broken object shapes. In contrast, our method learns object outlines well, thus recovering the local details better, \eg, the pear stem and the statue fingers. For 6 and 9 input views, more input views can provide more low-frequency pixel supervision for the rendering loss (\cref{eq:loss}). Thus, the performance of our method is similar to FreeNeRF. This also demonstrates the robustness of our method.

\begin{figure*}
    \centering
    \includegraphics[width=\linewidth]{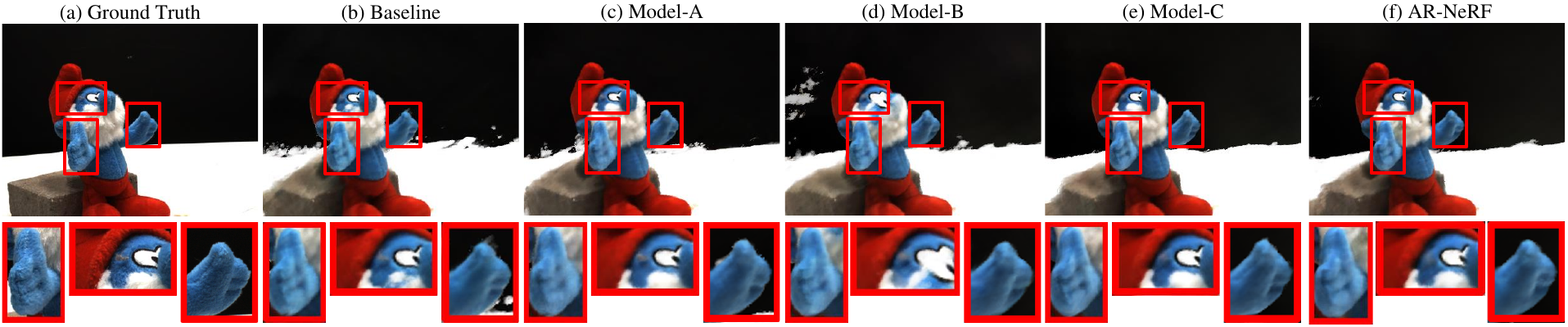}
    \caption{\textbf{Qualitative results of ablation study.}}
    \label{fig:ablation}
\end{figure*}

\customparagraph{LLFF dataset}  
\Cref{tab:llff} indicates that our method outperforms all methods in most metrics under different input-view settings. 
The pre-training methods trained on DTU degrade a lot on this dataset due to the domain gap. Despite fine-tuning on this dataset, they still cannot achieve satisfactory results. Among regularization methods, our method almost achieves the best performance. It is worth noting that our method outperforms other regularization methods by a considerable margin in 3 input-view setting. \cref{fig:llff_comp} shows the qualitative results in this setting. The compared regularization methods cannot render global structures well, \eg, the front of horns and the ceiling. In contrast, our method can synthesize these global structures well and learn fine details better, \eg, the pole and the bottom of cabinet. Note that, the scenes on LLFF are more complex than those on DTU. This results in the more complex frequency composition on LLFF than on DTU, posing great challenges to few-shot NeRF. Our AR-NeRF can adaptively align the frequency relationship between PE and image pixel supervision. Therefore, it has greater potential in handling the complex scenes on LLFF, making our performance improvement compared to FreeNeRF on LLFF is larger than that on DTU.

\subsection{Analysis}

In this section, we analyze the effectiveness of our designs on the DTU dataset under the 3 input-view setting. The batch size is set to 1024 for faster training.

\customparagraph{Ablation study} We adopt FreeNeRF~\cite{yang2023freenerf} as our baseline. It already incorporates the frequency regularization of PE and $\mathcal{L}_o$. Our different designs are progressively added to
the baseline to investigate their efficacy. The quantitative results are reported in \Cref{tab:ablation}. We see that, with the two-phase rendering supervision, Model-A improves PSNR a lot. This is because the two-phase rendering supervision makes the NeRF optimization focus more on learning low-frequency information. As shown in \cref{fig:ablation}(c), the global structures (\eg, the face) can be better rendered than Baseline (\cref{fig:ablation}(b)). With the adaptive rendering loss weight learning, all metrics of Model-B are improved compared to Baseline. \cref{fig:ablation}(d) shows that the local details, \eg, the synthesized hands become cleaner and clearer than Baseline and Model-A (\cref{fig:ablation}(b) and (c)). However, the adaptive rendering loss weight learning may be unstable in the early training phase, making the global structures of Model-B similar to Baseline. By combining these two designs, the performance of Model-C is further improved. \cref{fig:ablation}(e) shows that both global structures and local details are learned well. At last, since our proposed ray density regularization imposes sparsity constraints on the entire scene space to reduce noise, our method, AR-NeRF further improves PSNR and achieves the best results in all metrics.

\begin{table}[t]
    \centering
    \caption{
    \textbf{Ablation study.} The Baseline already incorporates the frequency regularization of PE and $\mathcal{L}_o$.  
    }
    {
\begin{tabular}{l|ccc|c|c|c|c}
\toprule
& $\mathcal{L}_s$ & $\mathcal{L}_u$ & $\mathcal{L}_r$ & {PSNR $\uparrow$} & {SSIM $\uparrow$} & {LPIPS $\downarrow$} & {Average $\downarrow$} \\
\midrule
Baseline & \multicolumn{3}{c|}{} & 19.54 & 0.766 & 0.221 & 0.111 \\
\midrule\midrule
Model-A & \checkmark & & & 19.81 & 0.767 & 0.222 & 0.109 \\
Model-B &  & \checkmark & & 19.73 & 0.771 & \bf 0.215 & 0.106 \\
Model-C & \checkmark & \checkmark & & 20.01 & \bf 0.773 & \bf 0.215 & 0.105 \\
AR-NeRF & \checkmark & \checkmark & \checkmark & \bf 20.07 & \bf 0.773 & \bf 0.215 & \bf 0.103 \\
\bottomrule\bottomrule
\end{tabular}}
    \label{tab:ablation}
\end{table}

\begin{wrapfigure}{r}{0.4\textwidth}
  \centering
  \includegraphics[width=0.4\textwidth]{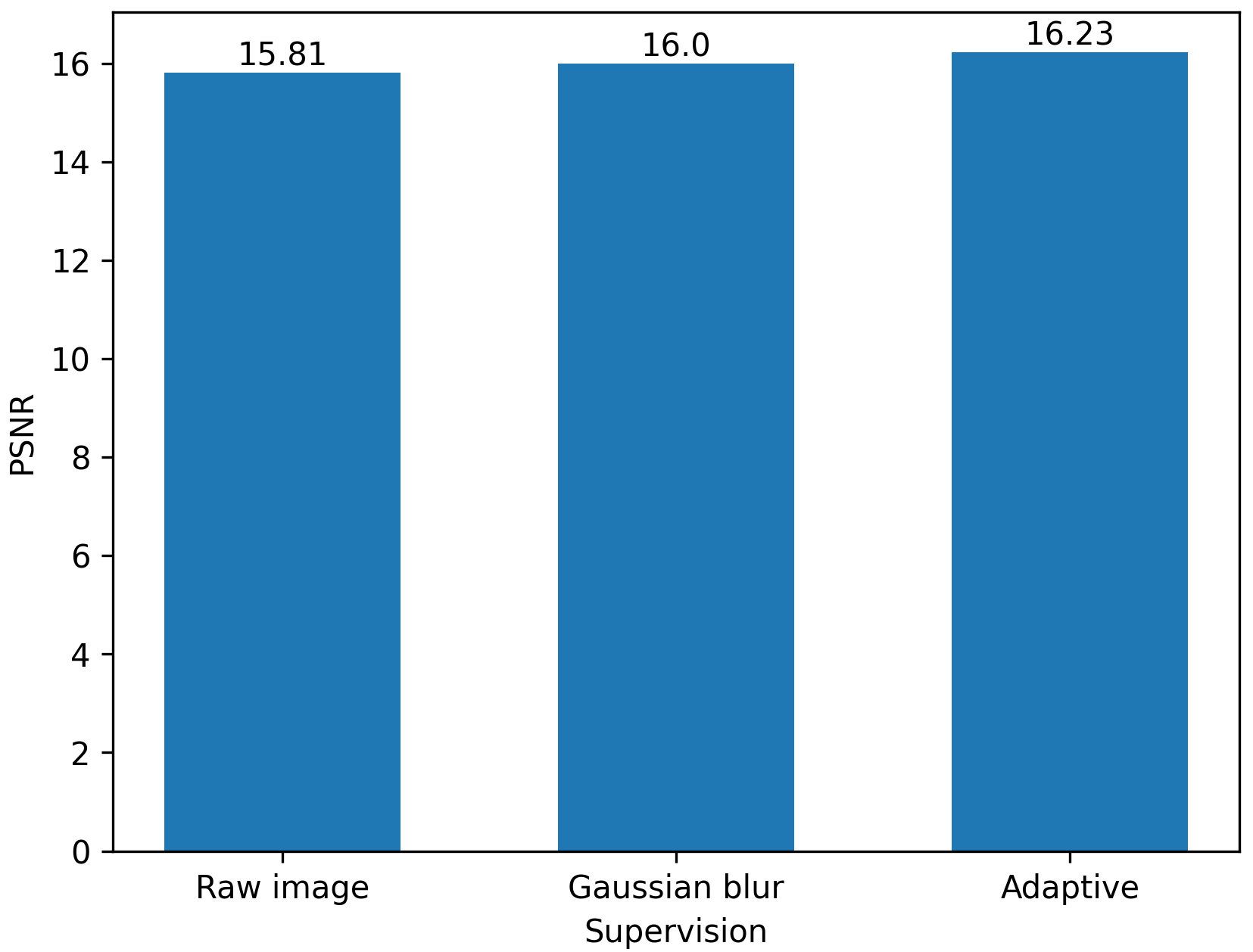} 
  \caption{\textbf{The impact of high-frequency supervision.} We compare the rendering quality of low-frequency pixels using different supervisions when only low-frequency PEs are enabled.} 
  \label{fig:hf_supervision}
\end{wrapfigure}

\customparagraph{The impact of high-frequency supervision} To investigate the impact of high-frequency supervision when learning low-frequency information, we only input low-frequency PEs. Specifically, only 10\% PEs are enabled for \cref{eq:pe}, and the following image pixel supervisions are studied: 1) raw image pixels, 2) applying Gaussian blur to raw image pixels, and 3) applying our adaptive rendering loss regularization.  
To assess the rendering quality of low-frequency image pixels, we leverage image edge information to distinguish low-frequency image pixels and only compute PSNR for these image pixels. The results in \cref{fig:hf_supervision} indicate that when only inputting low-frequency PEs, reducing high-frequency pixel supervision will help learn better low-frequency information. This greatly supports our motivation to align frequency relationship between PE and pixel supervision.

\begin{table}[t]
    \centering
    \caption{
    \textbf{Adaptive rendering loss regularization.} Linear means that the rendering loss weights for high-frequency pixel supervision are linearly increased from 0 to 1. 
    Adaptive Gaussian Blur means that we progressively reduce the standard variance of Gaussian blur kernel to construct adaptive blurred image supervision.  
    }
    {
\begin{tabular}{l|c|c|c|c}
\toprule
  &  PSNR $\uparrow$ & SSIM $\uparrow$ & LPIPS $\downarrow$ & Average $\downarrow$  \\
  \midrule
Baseline & 19.54 & 0.766 & 0.221 & 0.111 \\
Linear & 19.66 & 0.771 & 0.218 & 0.108 \\
Adaptive Gaussian Blur & 19.72 & 0.771 & 0.218 & 0.107 \\
Adaptive Rendering Loss (Ours) & \bf 20.01 & \bf 0.773 & \bf 0.215 & \bf 0.105 \\
\bottomrule
\end{tabular}}
    \label{tab:adaptive}
\end{table}

\customparagraph{Adaptive rendering loss regularization} To further investigate the impact of rendering loss regularization, we design a linear strategy and an adaptive Gaussian blur strategy for comparisons. For the linear strategy, based on the \cref{eq:loss}, we use image edge information to denote high-frequency pixels while other pixels are low-frequency pixels. Then, we introduce a rendering loss weight $h$ for \cref{eq:loss} as: $\mathcal{L}'=\mathcal{L}_\text{low} + h \cdot \mathcal{L}_\text{high}$. 
The weight $h$ is linearly increased from 0 to 1 for high-frequency pixel supervision during training. For the adaptive Gaussian blur strategy, based on the Model-A in \Cref{tab:ablation}, we progressively reduce the standard variance of Gaussian blur kernel to construct adaptive Gaussian blurred image supervision. As shown in \Cref{tab:adaptive}, both strategies improve the Baseline in all metrics. This also illustrates the inconsistency between the frequency regularization of PE and rendering loss. However, the linear and adaptive Gaussian blur strategies are handcrafted and cannot fully align the frequency relationship between PE and pixel supervision. In contrast, our design adaptively aligns the frequency relationship, thus achieving better results than the compared strategies.

\begin{table}[t]
    \centering
    \caption{
    \textbf{Ray density regularization.} Different choices of ray regularization are compared based on Model-C. 
    }
    {
\begin{tabular}{l|c|c|c|c}
\toprule
  &  PSNR $\uparrow$ & SSIM $\uparrow$ & LPIPS $\downarrow$ & Average $\downarrow$  \\
  \midrule
Model-C & 20.01 & 0.773 & 0.215 & 0.105 \\
Emptiness & 19.33 & 0.763 & 0.221 & 0.113 \\
Ray Density Entropy & 19.94 & 0.767 & 0.216 & 0.107 \\
Ray Density Reg. (Ours) & \bf 20.07 & \bf 0.773 & \bf 0.215 & \bf 0.103 \\
\bottomrule
\end{tabular}}
    \label{tab:ray}
\end{table}

\customparagraph{Ray density regularization} We further investigate the efficacy of ray density regularization. We adopt the Emptiness loss on $w_i$ in \cref{eq:render}~\cite{wang2023score} and the ray density entropy loss~\cite{kim2022infonerf} for comparison. As shown in \Cref{tab:ray}, both their results are worse than ours and the Model-C. For the Emptiness loss on $w_i$, we suspect this is because the $w_i$ in NeRF is a biased geometry proxy~\cite{wang2021neus,fu2022geo}. For the ray density entropy loss, it encourages unimodal distribution for ray densities, which is not necessarily true~\cite{wang2021neus,seo2023mixnerf}. Our proposed ray density regularization uses a relaxed geometry proxy and only imposes more penalties on small ray densities. Therefore, it can further improve the Model-C.
\section{Conclusion}
\label{sec:conclusion}

We have proposed AR-NeRF, an Adaptive Rendering loss regularization method to achieve high-quality novel view synthesis for few-shot NeRF. In our paper, we first reveal the inconsistency between the frequency regularization of PE and rendering loss. To mitigate this inconsistency, we propose adaptive rendering loss regularization, including two-phase rendering supervision and adaptive rendering loss weight learning. This effectively aligns the frequency relationship between PE and 2D-pixel supervision. Therefore, our AR-NeRF can learn better global structures in the early training phase, and adaptively learn local details throughout the training process. Experiments demonstrate that our method achieves state-of-the-art performance on both object-level and complex scenes.
\section*{Acknowledgements}

This research is supported by the National Research Foundation, Singapore under its AI Singapore Programme (AISG Award No: AISG2-RP-2021-022). Xuanyu Yi is supported by the Agency for Science, Technology AND Research, Singapore.

%
%
\bibliographystyle{splncs04}
\bibliography{main}

\begin{thebibliography}{10}
\providecommand{\url}[1]{\texttt{#1}}
\providecommand{\urlprefix}{URL }
\providecommand{\doi}[1]{https://doi.org/#1}

\bibitem{bai2023self}
Bai, J., Huang, L., Gong, W., Guo, J., Guo, Y.: Self-nerf: A self-training pipeline for few-shot neural radiance fields. arXiv preprint arXiv:2303.05775  (2023)

\bibitem{barron2021mip}
Barron, J.T., Mildenhall, B., Tancik, M., Hedman, P., Martin-Brualla, R., Srinivasan, P.P.: Mip-nerf: A multiscale representation for anti-aliasing neural radiance fields. In: Proceedings of the IEEE International Conference on Computer Vision. pp. 5855--5864 (2021)

\bibitem{barron2022mip}
Barron, J.T., Mildenhall, B., Verbin, D., Srinivasan, P.P., Hedman, P.: Mip-nerf 360: Unbounded anti-aliased neural radiance fields. In: Proceedings of the IEEE Conference on Computer Vision and Pattern Recognition. pp. 5470--5479 (2022)

\bibitem{bradbury2018jax}
Bradbury, J., Frostig, R., Hawkins, P., Johnson, M.J., Leary, C., Maclaurin, D., Necula, G., Paszke, A., VanderPlas, J., Wanderman-Milne, S., et~al.: Jax: composable transformations of python+ numpy programs. Version 0.2  \textbf{5},  14--24 (2018)

\bibitem{chan2021pi}
Chan, E.R., Monteiro, M., Kellnhofer, P., Wu, J., Wetzstein, G.: pi-gan: Periodic implicit generative adversarial networks for 3d-aware image synthesis. In: Proceedings of the IEEE conference on computer vision and pattern recognition. pp. 5799--5809 (2021)

\bibitem{chen2021mvsnerf}
Chen, A., Xu, Z., Zhao, F., Zhang, X., Xiang, F., Yu, J., Su, H.: Mvsnerf: Fast generalizable radiance field reconstruction from multi-view stereo. In: Proceedings of the IEEE International Conference on Computer Vision. pp. 14124--14133 (2021)

\bibitem{chen2022geoaug}
Chen, D., Liu, Y., Huang, L., Wang, B., Pan, P.: Geoaug: Data augmentation for few-shot nerf with geometry constraints. In: Proceedings of the European Conference on Computer Vision. pp. 322--337. Springer (2022)

\bibitem{chibane2021stereo}
Chibane, J., Bansal, A., Lazova, V., Pons-Moll, G.: Stereo radiance fields (srf): Learning view synthesis for sparse views of novel scenes. In: Proceedings of the IEEE Conference on Computer Vision and Pattern Recognition. pp. 7911--7920 (2021)

\bibitem{deng2022depth}
Deng, K., Liu, A., Zhu, J.Y., Ramanan, D.: Depth-supervised nerf: Fewer views and faster training for free. In: Proceedings of the IEEE Conference on Computer Vision and Pattern Recognition. pp. 12882--12891 (2022)

\bibitem{esposito2024geogen}
Esposito, S., Xu, Q., Kania, K., Hewitt, C., Mariotti, O., Petikam, L., Valentin, J., Onken, A., Mac~Aodha, O.: Geogen: Geometry-aware generative modeling via signed distance functions. In: Proceedings of the IEEE Conference on Computer Vision and Pattern Recognition. pp. 7479--7488 (2024)

\bibitem{fu2022geo}
Fu, Q., Xu, Q., Ong, Y.S., Tao, W.: Geo-neus: Geometry-consistent neural implicit surfaces learning for multi-view reconstruction. Advances in Neural Information Processing Systems  \textbf{35},  3403--3416 (2022)

\bibitem{jain2021putting}
Jain, A., Tancik, M., Abbeel, P.: Putting nerf on a diet: Semantically consistent few-shot view synthesis. In: Proceedings of the IEEE International Conference on Computer Vision. pp. 5885--5894 (2021)

\bibitem{jensen2014large}
Jensen, R., Dahl, A., Vogiatzis, G., Tola, E., Aan{\ae}s, H.: Large scale multi-view stereopsis evaluation. In: Proceedings of the IEEE Conference on Computer Vision and Pattern Recognition. pp. 406--413 (2014)

\bibitem{johari2022geonerf}
Johari, M.M., Lepoittevin, Y., Fleuret, F.: Geonerf: Generalizing nerf with geometry priors. In: Proceedings of the IEEE Conference on Computer Vision and Pattern Recognition. pp. 18365--18375 (2022)

\bibitem{kendall2017uncertainties}
Kendall, A., Gal, Y.: What uncertainties do we need in bayesian deep learning for computer vision? Advances in neural information processing systems  \textbf{30} (2017)

\bibitem{kim2022infonerf}
Kim, M., Seo, S., Han, B.: Infonerf: Ray entropy minimization for few-shot neural volume rendering. In: Proceedings of the IEEE Conference on Computer Vision and Pattern Recognition. pp. 12912--12921 (2022)

\bibitem{kingma2014adam}
Kingma, D.P., Ba, J.: Adam: A method for stochastic optimization. arXiv preprint arXiv:1412.6980  (2014)

\bibitem{kwak2023geconerf}
Kwak, M., Song, J., Kim, S.: Geconerf: Few-shot neural radiance fields via geometric consistency. arXiv preprint arXiv:2301.10941  (2023)

\bibitem{lin2023magic3d}
Lin, C.H., Gao, J., Tang, L., Takikawa, T., Zeng, X., Huang, X., Kreis, K., Fidler, S., Liu, M.Y., Lin, T.Y.: Magic3d: High-resolution text-to-3d content creation. In: Proceedings of the IEEE Conference on Computer Vision and Pattern Recognition. pp. 300--309 (2023)

\bibitem{martin2021nerf}
Martin-Brualla, R., Radwan, N., Sajjadi, M.S., Barron, J.T., Dosovitskiy, A., Duckworth, D.: Nerf in the wild: Neural radiance fields for unconstrained photo collections. In: Proceedings of the IEEE Conference on Computer Vision and Pattern Recognition. pp. 7210--7219 (2021)

\bibitem{max1995optical}
Max, N.: Optical models for direct volume rendering. IEEE Transactions on Visualization and Computer Graphics  \textbf{1}(2),  99--108 (1995)

\bibitem{mildenhall2019local}
Mildenhall, B., Srinivasan, P.P., Ortiz-Cayon, R., Kalantari, N.K., Ramamoorthi, R., Ng, R., Kar, A.: Local light field fusion: Practical view synthesis with prescriptive sampling guidelines. ACM Transactions on Graphics (TOG)  \textbf{38}(4),  1--14 (2019)

\bibitem{mildenhall2021nerf}
Mildenhall, B., Srinivasan, P.P., Tancik, M., Barron, J.T., Ramamoorthi, R., Ng, R.: Nerf: Representing scenes as neural radiance fields for view synthesis. Communications of the ACM  \textbf{65}(1),  99--106 (2021)

\bibitem{niemeyer2022regnerf}
Niemeyer, M., Barron, J.T., Mildenhall, B., Sajjadi, M.S., Geiger, A., Radwan, N.: Regnerf: Regularizing neural radiance fields for view synthesis from sparse inputs. In: Proceedings of the IEEE Conference on Computer Vision and Pattern Recognition. pp. 5480--5490 (2022)

\bibitem{pan2022activenerf}
Pan, X., Lai, Z., Song, S., Huang, G.: Activenerf: Learning where to see with uncertainty estimation. In: Proceedings of the European Conference on Computer Vision. pp. 230--246. Springer (2022)

\bibitem{poole2022dreamfusion}
Poole, B., Jain, A., Barron, J.T., Mildenhall, B.: Dreamfusion: Text-to-3d using 2d diffusion. arXiv preprint arXiv:2209.14988  (2022)

\bibitem{ren2023hierarchical}
Ren, C., Xu, Q., Zhang, S., Yang, J.: Hierarchical prior mining for non-local multi-view stereo. In: Proceedings of the IEEE International Conference on Computer Vision. pp. 3611--3620 (2023)

\bibitem{roessle2022dense}
Roessle, B., Barron, J.T., Mildenhall, B., Srinivasan, P.P., Nie{\ss}ner, M.: Dense depth priors for neural radiance fields from sparse input views. In: Proceedings of the IEEE Conference on Computer Vision and Pattern Recognition. pp. 12892--12901 (2022)

\bibitem{seo2023flipnerf}
Seo, S., Chang, Y., Kwak, N.: Flipnerf: Flipped reflection rays for few-shot novel view synthesis. In: Proceedings of the IEEE International Conference on Computer Vision. pp. 22883--22893 (2023)

\bibitem{seo2023mixnerf}
Seo, S., Han, D., Chang, Y., Kwak, N.: Mixnerf: Modeling a ray with mixture density for novel view synthesis from sparse inputs. In: Proceedings of the IEEE Conference on Computer Vision and Pattern Recognition. pp. 20659--20668 (2023)

\bibitem{shen2022conditional}
Shen, J., Agudo, A., Moreno-Noguer, F., Ruiz, A.: Conditional-flow nerf: Accurate 3d modelling with reliable uncertainty quantification. In: Proceedings of the European Conference on Computer Vision. pp. 540--557. Springer (2022)

\bibitem{shen2021stochastic}
Shen, J., Ruiz, A., Agudo, A., Moreno-Noguer, F.: Stochastic neural radiance fields: Quantifying uncertainty in implicit 3d representations. In: Proceedings of the International Conference on 3D Vision. pp. 972--981. IEEE (2021)

\bibitem{su2024psdf}
Su, W., Zhang, C., Xu, Q., Tao, W.: Psdf: Prior-driven neural implicit surface learning for multi-view reconstruction. arXiv preprint arXiv:2401.12751  (2024)

\bibitem{sun2024global}
Sun, X., Xu, Q., Yang, X., Zang, Y., Wang, C.: Global and hierarchical geometry consistency priors for few-shot nerfs in indoor scenes. In: Proceedings of the IEEE Conference on Computer Vision and Pattern Recognition. pp. 20530--20539 (2024)

\bibitem{tancik2020fourier}
Tancik, M., Srinivasan, P., Mildenhall, B., Fridovich-Keil, S., Raghavan, N., Singhal, U., Ramamoorthi, R., Barron, J., Ng, R.: Fourier features let networks learn high frequency functions in low dimensional domains. Advances in Neural Information Processing Systems  \textbf{33},  7537--7547 (2020)

\bibitem{truong2023sparf}
Truong, P., Rakotosaona, M.J., Manhardt, F., Tombari, F.: Sparf: Neural radiance fields from sparse and noisy poses. In: Proceedings of the IEEE Conference on Computer Vision and Pattern Recognition. pp. 4190--4200 (2023)

\bibitem{uy2023scade}
Uy, M.A., Martin-Brualla, R., Guibas, L., Li, K.: Scade: Nerfs from space carving with ambiguity-aware depth estimates. In: Proceedings of the IEEE Conference on Computer Vision and Pattern Recognition. pp. 16518--16527 (2023)

\bibitem{verbin2022ref}
Verbin, D., Hedman, P., Mildenhall, B., Zickler, T., Barron, J.T., Srinivasan, P.P.: Ref-nerf: Structured view-dependent appearance for neural radiance fields. In: 2022 IEEE Conference on Computer Vision and Pattern Recognition. pp. 5481--5490. IEEE (2022)

\bibitem{wang2023sparsenerf}
Wang, G., Chen, Z., Loy, C.C., Liu, Z.: Sparsenerf: Distilling depth ranking for few-shot novel view synthesis. arXiv preprint arXiv:2303.16196  (2023)

\bibitem{wang2024pgahum}
Wang, H., Xu, Q., Chen, H., Ma, R.: Pgahum: Prior-guided geometry and appearance learning for high-fidelity animatable human reconstruction. arXiv preprint arXiv:2404.13862  (2024)

\bibitem{wang2023score}
Wang, H., Du, X., Li, J., Yeh, R.A., Shakhnarovich, G.: Score jacobian chaining: Lifting pretrained 2d diffusion models for 3d generation. In: Proceedings of the IEEE Conference on Computer Vision and Pattern Recognition. pp. 12619--12629 (2023)

\bibitem{wang2021neus}
Wang, P., Liu, L., Liu, Y., Theobalt, C., Komura, T., Wang, W.: Neus: Learning neural implicit surfaces by volume rendering for multi-view reconstruction. Advances in Neural Information Processing Systems  (2021)

\bibitem{wang2004image}
Wang, Z., Bovik, A.C., Sheikh, H.R., Simoncelli, E.P.: Image quality assessment: from error visibility to structural similarity. IEEE transactions on image processing  \textbf{13}(4),  600--612 (2004)

\bibitem{wu2024reconfusion}
Wu, R., Mildenhall, B., Henzler, P., Park, K., Gao, R., Watson, D., Srinivasan, P.P., Verbin, D., Barron, J.T., Poole, B., et~al.: Reconfusion: 3d reconstruction with diffusion priors. In: Proceedings of the IEEE Conference on Computer Vision and Pattern Recognition. pp. 21551--21561 (2024)

\bibitem{wynn2023diffusionerf}
Wynn, J., Turmukhambetov, D.: Diffusionerf: Regularizing neural radiance fields with denoising diffusion models. In: Proceedings of the IEEE Conference on Computer Vision and Pattern Recognition. pp. 4180--4189 (2023)

\bibitem{xu2023pr}
Xu, J., Xu, Q., Liao, X., Su, W., Zhang, C., Ong, Y.S., Tao, W.: Pr-neus: A prior-based residual learning paradigm for fast multi-view neural surface reconstruction. arXiv preprint arXiv:2312.11577  (2023)

\bibitem{xu2022multi}
Xu, Q., Kong, W., Tao, W., Pollefeys, M.: Multi-scale geometric consistency guided and planar prior assisted multi-view stereo. IEEE Transactions on Pattern Analysis and Machine Intelligence  \textbf{45}(4),  4945--4963 (2022)

\bibitem{xu2019multi}
Xu, Q., Tao, W.: Multi-scale geometric consistency guided multi-view stereo. In: Proceedings of the IEEE conference on computer vision and pattern recognition. pp. 5483--5492 (2019)

\bibitem{yang2023freenerf}
Yang, J., Pavone, M., Wang, Y.: Freenerf: Improving few-shot neural rendering with free frequency regularization. In: Proceedings of the IEEE Conference on Computer Vision and Pattern Recognition. pp. 8254--8263 (2023)

\bibitem{yariv2021volume}
Yariv, L., Gu, J., Kasten, Y., Lipman, Y.: Volume rendering of neural implicit surfaces. Advances in Neural Information Processing Systems  \textbf{34},  4805--4815 (2021)

\bibitem{yi2024diffusion}
Yi, X., Wu, Z., Xu, Q., Zhou, P., Lim, J.H., Zhang, H.: Diffusion time-step curriculum for one image to 3d generation. In: Proceedings of the IEEE Conference on Computer Vision and Pattern Recognition. pp. 9948--9958 (2024)

\bibitem{yu2021pixelnerf}
Yu, A., Ye, V., Tancik, M., Kanazawa, A.: pixelnerf: Neural radiance fields from one or few images. In: Proceedings of the IEEE Conference on Computer Vision and Pattern Recognition. pp. 4578--4587 (2021)

\bibitem{zhang2018unreasonable}
Zhang, R., Isola, P., Efros, A.A., Shechtman, E., Wang, O.: The unreasonable effectiveness of deep features as a perceptual metric. In: Proceedings of the IEEE conference on computer vision and pattern recognition. pp. 586--595 (2018)

\end{thebibliography}

\end{document}